\documentclass[10pt,twocolumn,letterpaper]{article}

\usepackage{cvpr}
\usepackage{booktabs}
\usepackage{times}
\usepackage{epsfig}
\usepackage{graphicx}
\usepackage{amsmath}
\usepackage{amssymb}
\usepackage{gensymb}
\usepackage{booktabs}
\usepackage{graphicx}
\usepackage[table,xcdraw]{xcolor}
\usepackage{array}
\usepackage{tabularx}
\usepackage{subfigure}
\usepackage{blindtext}
\usepackage{enumitem}
\usepackage{colortbl}
\usepackage{xcolor}
\usepackage{multirow}
\usepackage{graphicx}
\usepackage[table,xcdraw]{xcolor}
\usepackage[normalem]{ulem}

\newcolumntype{C}[1]{>{\centering\arraybackslash}m{#1}}

\usepackage[pagebackref=true,breaklinks=true,letterpaper=true,colorlinks,bookmarks=false]{hyperref}

\cvprfinalcopy 

\begin{document}

\title{Deep Shape from Polarization}

\author{Yunhao Ba$^1$ \quad Alex Gilbert$^{1*}$ \quad Franklin Wang$^{1}$\thanks{: Authors contributed equally}  \quad Jinfa Yang$^2$ \quad Rui Chen$^2$ \quad Yiqin Wang$^1$ \\ \quad Lei Yan$^2$ \quad Boxin Shi$^2$  \quad Achuta Kadambi$^1$ \vspace{0.3em}\\
$^1$University of California, Los Angeles \quad$^2$Peking University \\
}
\maketitle

\newcommand{\SubSection}[1]{\vspace{-.00cm}\textcolor{black}{\subsection{#1}}\vspace{-.00cm}}
\newcommand{\Section}[1]{\vspace{-.00cm}\textcolor{black}{\section{#1}}\vspace{-.00cm}}
\newcommand{\SubSectionL}[2]{\vspace{-.00cm}\textcolor{black}{\subsection{#1}\label{#2}}\vspace{-.00cm}}

\begin{abstract}
This paper makes a first attempt to bring the Shape from Polarization (SfP) problem to the realm of deep learning. The previous state-of-the-art methods for SfP have been purely physics-based. We see value in these principled models, and blend these physical models as priors into a neural network architecture. This proposed approach achieves results that exceed the previous state-of-the-art on a challenging dataset we introduce. This dataset consists of polarization images taken over a range of object textures, paints, and lighting conditions. We report that our proposed method achieves the lowest test error on each tested condition in our dataset, showing the value of blending data-driven and physics-driven approaches. 
\end{abstract}


\section{Introduction}

While deep learning has revolutionized many areas of computer vision, the deep learning revolution has not yet been studied in context of Shape from Polarization (SfP). The SfP problem is fascinating because, if successful, shape could be obtained in completely passive lighting conditions without estimating lighting direction. Recent progress in CMOS sensors has spawned machine vision cameras that capture the required polarization information in a single shot~\cite{PolarM}, making the capture process more relaxed than photometric stereo.

This SfP problem can be stated simply: light that reflects off an object has a polarization state that corresponds to shape. In reality, the underlying physics is among the most optically complex of all computer vision problems. For this reason, previous SfP methods have high error rates (in context of mean angular error (MAE) of surface normal estimation), and limited generalization to mixed materials and lighting conditions. 

The physics of SfP are based on the Fresnel Equations. These equations lead to an underdetermined system, in light of the so-called \emph{ambiguity problem}. This problem arises because a linear polarizer cannot distinguish between polarized light that is rotated by $\pi$ radians. This results in two confounding estimates for azimuth angle at each pixel. Previous work in SfP has used additional information to constrain the ambiguity problem. For instance, Smith \textit{et al.}~\cite{smith2016linear} use both polarization and shading constraints as linear equations when solving object depth, and Mahmoud \textit{et al.}~\cite{mahmoud2012direct} use shape from shading constraints to correct the ambiguities. Other authors assume surface convexity to constrain the azimuth angle~\cite{miyazaki2003polarization, atkinson2006recovery}. Yet another solution is to use a coarse depth map to constrain the ambiguity~\cite{kadambi2015polarized, Kadambi17}. There are also additional binary ambiguities based on reflection type, as discussed in~\cite{atkinson2006recovery,mahmoud2012direct}. Table~\ref{tab:pos} compares tradeoffs of our proposed technique in context of prior work.

\begin{table}
\caption{\textbf{The proposed hybrid of physics and learning compared to previous SfP methods, which are physics-based.}}
\vspace{-.5cm}
\begin{center}
\resizebox{.48\textwidth}{!}{%
\begin{tabular}{@{}ccccc@{}}\setlength{\textfloatsep}{.1cm}

 \cellcolor[HTML]{656565}\rule{0pt}{0ex}\textcolor{white}{\textbf{Method}} & \cellcolor[HTML]{656565}\rule{0pt}{0ex}\textcolor{white}{\textbf{Inputs}} & \cellcolor[HTML]{656565}\begin{tabular}[c]{@{}c@{}}\textcolor{white}{\textbf{Mean Angular}} \\ \textcolor{white}{\textbf{Error}} \end{tabular} & \cellcolor[HTML]{656565}\begin{tabular}[c]{@{}c@{}}\textcolor{white}{\textbf{Robustness to}}\\ \textcolor{white}{\textbf{Texture-Copy}}\end{tabular} & \cellcolor[HTML]{656565}\begin{tabular}[c]{@{}c@{}}\textcolor{white}{\textbf{Lighting}} \\ \textcolor{white}{\textbf{Invariance}}\end{tabular} \\[1ex]
\cellcolor[HTML]{EFEFEF}\begin{tabular}[c]{@{}c@{}}\rule{0pt}{3ex}Miyazaki \cite{miyazaki2003polarization}\end{tabular} & \cellcolor[HTML]{9AFF99}\begin{tabular}[c]{@{}c@{}}\rule{0pt}{3ex}Polar. Images\end{tabular} & \cellcolor[HTML]{FFCCC9}\begin{tabular}[c]{@{}c@{}}\rule{0pt}{3ex}High\end{tabular} & \cellcolor[HTML]{9AFF99}\begin{tabular}[c]{@{}c@{}}\rule{0pt}{3ex}Strong\end{tabular} & \cellcolor[HTML]{FFFFC7}\begin{tabular}[c]{@{}c@{}}\rule{0pt}{3ex}Moderate\end{tabular} \\[1ex]
\cellcolor[HTML]{EFEFEF}\begin{tabular}[c]{@{}c@{}}\rule{0pt}{3ex} \centering Mahmoud \cite{mahmoud2012direct} \end{tabular} & \cellcolor[HTML]{9AFF99}\begin{tabular}[c]{@{}c@{}}\rule{0pt}{3ex} Polar. Images\end{tabular} & \cellcolor[HTML]{FFCCC9}\begin{tabular}[c]{@{}c@{}}\rule{0pt}{3ex}High\end{tabular} & \cellcolor[HTML]{FFCCC9}\begin{tabular}[c]{@{}c@{}}\rule{0pt}{3ex}Not Observed\end{tabular} & \cellcolor[HTML]{FFFFC7}\begin{tabular}[c]{@{}c@{}}\rule{0pt}{3ex}Moderate\end{tabular} \\[1ex]
 
\cellcolor[HTML]{EFEFEF}\begin{tabular}{c}\rule{0pt}{0ex} Smith~\cite{Smith18} \end{tabular} & \cellcolor[HTML]{FFCCC9}\begin{tabular}[c]{@{}c@{}}\rule{0pt}{0ex}Polar. Images\\ Lighting est.\end{tabular} & \cellcolor[HTML]{FFFFC7}Moderate & \cellcolor[HTML]{9AFF99}Strong & \cellcolor[HTML]{FFFFC7}Moderate\\[1ex]

\cellcolor[HTML]{EFEFEF}\rule{0pt}{3ex}Proposed & \cellcolor[HTML]{9AFF99}\rule{0pt}{3ex}Polar. Images & \cellcolor[HTML]{9AFF99}\rule{0pt}{3ex}Lowest & \cellcolor[HTML]{9AFF99}\rule{0pt}{3ex}Strong & \cellcolor[HTML]{9AFF99}\rule{0pt}{3ex}Strong\\[1ex]

\end{tabular}%
} 
\end{center}

    \label{tab:pos}

\end{table}

Another contributing factor to the underdetermined nature of SfP is the \emph{refractive problem}. SfP needs knowledge of per-pixel refractive indices. Previous work has used hard-coded values to estimate the refractive index of scenes~\cite{miyazaki2003polarization}. This leads to a relative shape recovered with refractive distortion. 

Yet another limitation of the physical model is particular susceptibility to \emph{noise}. The polarization signal is very subtle for fronto-parallel geometries so it is important that  the input images are relatively noise-free. Unfortunately, a polarizing filter reduces the captured light intensity by 50 percent, worsening the effects of Poisson shot noise, encouraging a noise tolerant SfP algorithm.\footnote{For a detailed discussion of other sources of noise, such as rotation bias, please refer to Schechner~\cite{schechner2015self}.} 

In this paper, we address these SfP pitfalls by moving away from a physics-only solution, toward the realm of data-driven techniques. While it is tempting to apply traditional deep learning models to the SfP problem, we find this approach does not maximize performance. Instead, we propose a physics-based learning algorithm that not only outperforms traditional deep learning, but also outperforms three baseline comparisons to physics-based SfP. We summarize our contributions as follows:
\vspace{.2cm}
\begin{itemize}[itemsep=0.8em]
    \item[$\bullet$] a first attempt to apply deep learning techniques to solve the SfP problem;
    \item[$\bullet$] incorporation of the existing physical model into the deep learning approach;  
    \item[$\bullet$] demonstration of significant error reduction; and 
    \item[$\bullet$] introduction of the first polarization image dataset with ground truth shape, laying a foundation for future data-driven methods. 
\end{itemize}
\vspace{.2cm}

\noindent\textbf{Limitations:} As a physics-based learning approach, our technique still relies on computing the physical priors for every test example. This means that the per-frame runtime would be the sum of the compute time for the forward pass and that of the physics-based prior. Future work could parallelize compute of the physical prior, to achieve real-time performance. Another limitation pertains to the accuracy inherent to SfP. Our average MAE on the test set is 18.5 degrees. While this is the best SfP performer on our challenging dataset, the error is higher than what one can obtain with a more controlled technique like photometric stereo.

\section{Related Work}

Polarization cues have been employed previously for different tasks, such as reflectometry estimation \cite{ghosh2010circularly}, radiometric calibration~\cite{Teo_2018_CVPR}, facial geometry reconstruction~\cite{ghosh2011multiview}, dynamic interferometry~\cite{maeda2018dynamic}, polarimetric spatially varying surface reflectance functions (SVBRDF) recovery~\cite{baek2018simultaneous}, and object shape acquisition~\cite{ma2007rapid, guarnera2012estimating, riviere2017polarization,Zhu_2019_CVPR}. This paper sits at the seamline of deep learning and SfP, offering unique performance tradeoffs from prior work. Refer to Table~\ref{tab:pos} for an overview. \newline
\newline

\noindent\textbf{Shape from Polarization\,\,} infers the shape (usually represented in surface normals) of a surface by observing the correlated changes of image intensity with the polarization information. Changes of polarization information could be captured by rotating a linear polarizer in front of an ordinary camera~\cite{Wolff97, Atkinson18} or polarization cameras using a single shot in real time (e.g., PolarM~\cite{PolarM} camera used in~\cite{Yang18}). Conventional Shape from Polarization decodes such information to recover the surface normal up to some ambiguity. If only images with different polarization information are available, heuristic priors such as the surface normals along the boundary and convexity of the objects are employed to remove the ambiguity~\cite{miyazaki2003polarization, atkinson2006recovery}. Photometric constraints from shape from shading~\cite{mahmoud2012direct} and photometric stereo~\cite{Drbohlav01, Ngo15, Atkinson17} complements  polarization constraints to make the normal estimates unique. If multi-spectral measurements are available, surface normal and its refractive index could be estimated at the same time~\cite{Huynh10, Huynh13}. More recently, a joint formulation of shape from shading and shape from polarization in a linear manner is shown to be able to directly estimate the depth of the surface~\cite{smith2016linear, Tozza17, Smith18}. This paper is the first attempt at studying deep learning and SfP together. \newline

\noindent
\textbf{Polarized 3D\,\,} involves stronger assumptions than SfP and has different inputs and outputs. Recognizing that SfP alone is a limited technique, the Polarized 3D class of methods integrate shape from polarization with a low resolution depth estimate. This additional constraint allows not just recovery of shape but also a high-quality 3D model. The low resolution depth could be achieved by employing two-view~\cite{Miyazaki04,Atkinson05,Berger17}, three-view~\cite{Chen18}, multi-view~\cite{Miyazaki16,Cui17} stereo, or even in real time by using a SLAM system~\cite{Yang18}. These depth estimates from geometric methods are not reliable in textureless regions where finding correspondence for triangulation is difficult. Polarimetric cues could be jointly used to improve such unreliable depth estimates to obtain a more complete shape estimation. A depth sensor such as the Kinect can also provide coarse depth prior to disambiguate the ambiguous normal estimates given by SfP~\cite{kadambi2015polarized,Kadambi17}. The key step that characterizes Polarized 3D is a holistic approach that rethinks both SfP and the depth-normal fusion process. The main limitation of Polarized 3D is the strong requirement of a coarse depth map, which is not true for our proposed technique. \newline
\newline
\noindent
\textbf{Data-driven computational imaging\,\,}  approaches draw much attention in recent years thanks to the powerful modeling ability of deep neural networks. Various types of convolutional neural networks (CNNs) are designed to enable 3D imaging for many types of sensors and measurements. From single photon sensor measurements, a multi-scale denoising and upsampling CNN is proposed to refine depth estimates~\cite{Lindell18}.  CNNs also show advantage in solving phase unwrapping, multipath interference, and denoising jointly from raw time-of-flight measurements~\cite{marco2017deeptof,Su18}. From multi-directional lighting measurements, a fully-connected network is proposed to solve photometric stereo for general reflectance with a pre-defined set of light directions~\cite{Santo17}. Then the fully-convolutional network with an order-agnostic max-pooling operation~\cite{Chen18PS} and the observation map invariant to the number and permutation of the images~\cite{Ikehata18} are concurrently proposed to deal with an arbitrary set of light directions. Normal estimates from photometric stereo can also be learned in an unsupervised manner by minimizing reconstruction loss~\cite{Taniai18}. Other than 3D imaging, deep learning has helped solve several inverse problems in the field of computational imaging~\cite{satat2017object, tancik2018flash, tancik2018data}. Separation of shape, reflectance and illuminance maps for wild facial images can be achieved with the CNNs as well~\cite{sengupta2018sfsnet}. CNNs also exhibit potential for modeling SVBRDF of a near-planar surface~\cite{li2017modeling, ye2018single, li2018materials, deschaintre2018single}, and more complex objects~\cite{li2018learning_sv}. The challenge with existing deep learning frameworks is that they do not leverage the unique physics of polarization.

\section{Proposed Method}

In this section, we first introduce basic knowledge of SfP, and then present our physics-based CNN. Blending physics and deep learning improves the performance and generalizability of the method. 

\subsection{Image formation and physical solution}\label{physical_solutions}

Our objective is to reconstruct surface normals $\boldsymbol{\hat{N}}$ from a set of polarization images \{$\boldsymbol{I}_{\phi_{1}}$, $\boldsymbol{I}_{\phi_{2}}$, ..., $\boldsymbol{I}_{\phi_{M}}$\} with different polarization angles. For a specific polarization angle $\phi_{pol}$, the intensity at a pixel of a captured image follows a sinusoidal variation under unpolarized illumination:
\begin{equation}
  I(\phi_{pol}) = \frac{I_{max} + I_{min}}{2} + \frac{I_{max} - I_{min}}{2}\cos(2(\phi_{pol} - \phi)),
\label{eq:TRS}
\end{equation}
where $\phi$ denotes the phase angle, and $I_{min}$ and $I_{max}$ are lower and upper bounds for the observed intensity. Equation (\ref{eq:TRS}) has a $\pi$-\emph{\textbf{ambiguity}} in context of $\phi$: two phase angles, with a $\pi$ shift, will result in the same intensity in the captured images. Based on the phase angle $\phi$, the azimuth angle $\varphi$ can be retrieved with $\frac{\pi}{2}$-\emph{\textbf{ambiguity}} as follows~\cite{Cui17}:
\begin{equation}
  \phi = \begin{cases}
    \varphi, & \text{if diffuse reflection dominates}\\
    \varphi - \frac{\pi}{2}, & \text{if specular reflection dominates}
  \end{cases}.
\end{equation}

The zenith angle $\theta$ is related to the degree of polarization $\rho$, which can be written as:
\begin{equation}
  \rho = \frac{I_{max} - I_{min}}{I_{max} + I_{min}}.
\label{eq:int_to_dop}
\end{equation}

When diffuse reflection is dominant, the degree of polarization can be expressed with the zenith angle $\theta$ and the refractive index $n$ as follows~\cite{atkinson2006recovery}:
\begin{equation} \label{diffuse rho}
  \rho_{d} = \frac{(n - \frac{1}{n})^2 \sin^2\theta}{2 + 2n^2 - (n + \frac{1}{n})^2 \sin^2\theta + 4\cos\theta \sqrt{n^2 - \sin^2\theta}}.
\end{equation}

The dependency of $\rho_{d}$ on $n$ is weak~\cite{atkinson2006recovery}, and we assume $n = 1.5$ throughout the rest of this paper. With this known $n$, Equation (\ref{diffuse rho}) can be rearranged to obtain a close-form estimation of the zenith angle for the diffuse dominant case. 

\begin{figure}[t]
  \centering
  \includegraphics[width=\columnwidth]{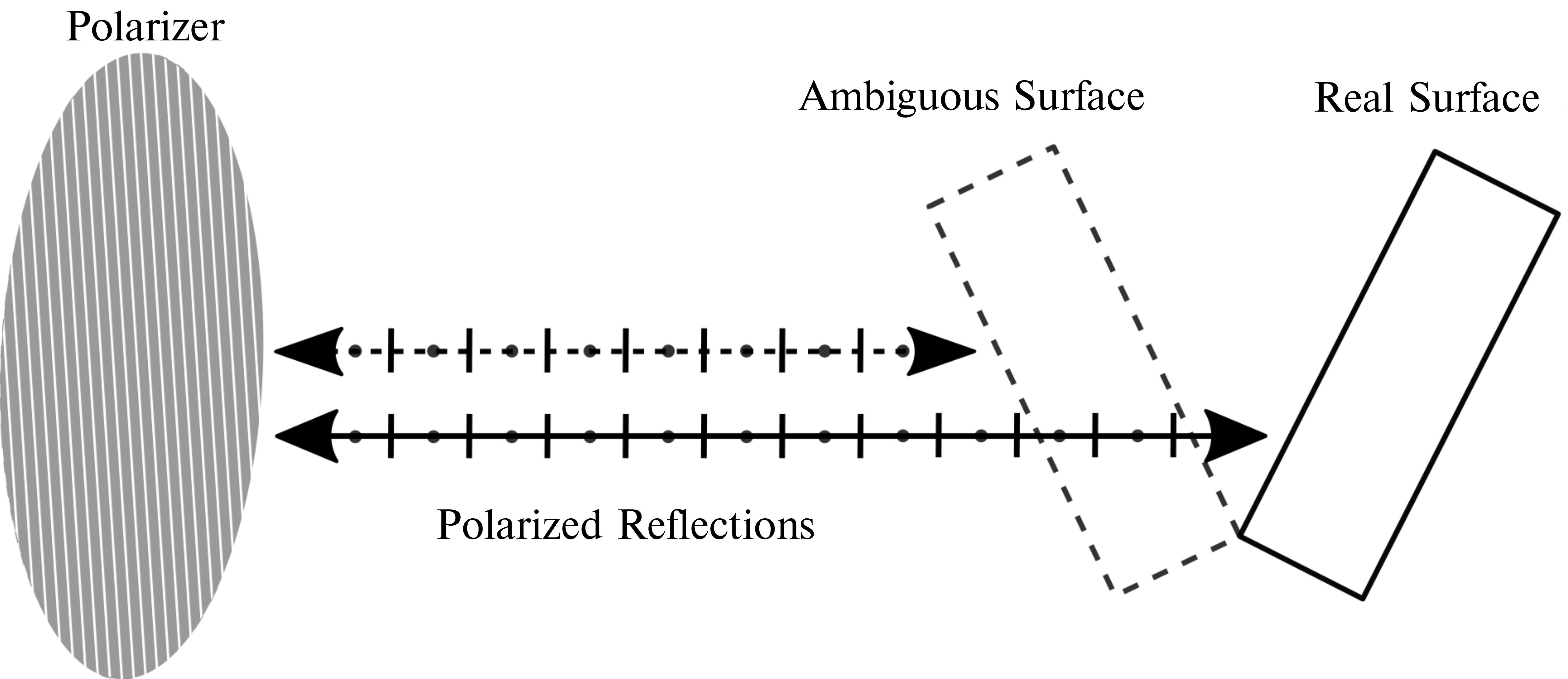}
    \caption{\textbf{SfP is underdetermined and one causal factor is the \emph{ambiguity problem}.} Here, two different surface orientations could result in exactly the same polarization signal, represented by dots and hashes. The dots represent polarization out of the plane of the paper and the hashes represent polarization within the plane of the board. Based on the measured data, it is unclear which orientation is correct. Ambiguities can also arise due to specular and diffuse reflections (which change the phase of light). For this reason, our network uses multiple physical priors.}
    \label{fig:amb}
\end{figure}

When specular reflection is dominant, the degree of polarization can be written as~\cite{atkinson2006recovery}:
 \begin{equation} \label{specualr rho}
   \rho_{s} = \frac{2 \sin^2\theta \cos\theta \sqrt{n^2 - \sin^2\theta}}{n^2 - \sin^2\theta - n^2\sin^2\theta + 2\sin^4\theta}.
 \end{equation}
Equation (\ref{specualr rho}) can not be inverted analytically, and solving the zenith angle with numerical interpolation will produce two solutions if there are no additional constraints. For real world objects, specular reflection and diffuse reflection are mixed depending on the surface material of the object. As shown in Figure~\ref{fig:amb}, the ambiguity in the azimuth angle and uncertainty in the zenith angle are fundamental limitations of SfP. Overcoming these limitations through physics-based neural networks is the primary focus of this paper. 

\subsection{Learning with physics}\label{learning_with_physics}

\begin{figure*}[t]
  \includegraphics[width=\textwidth]{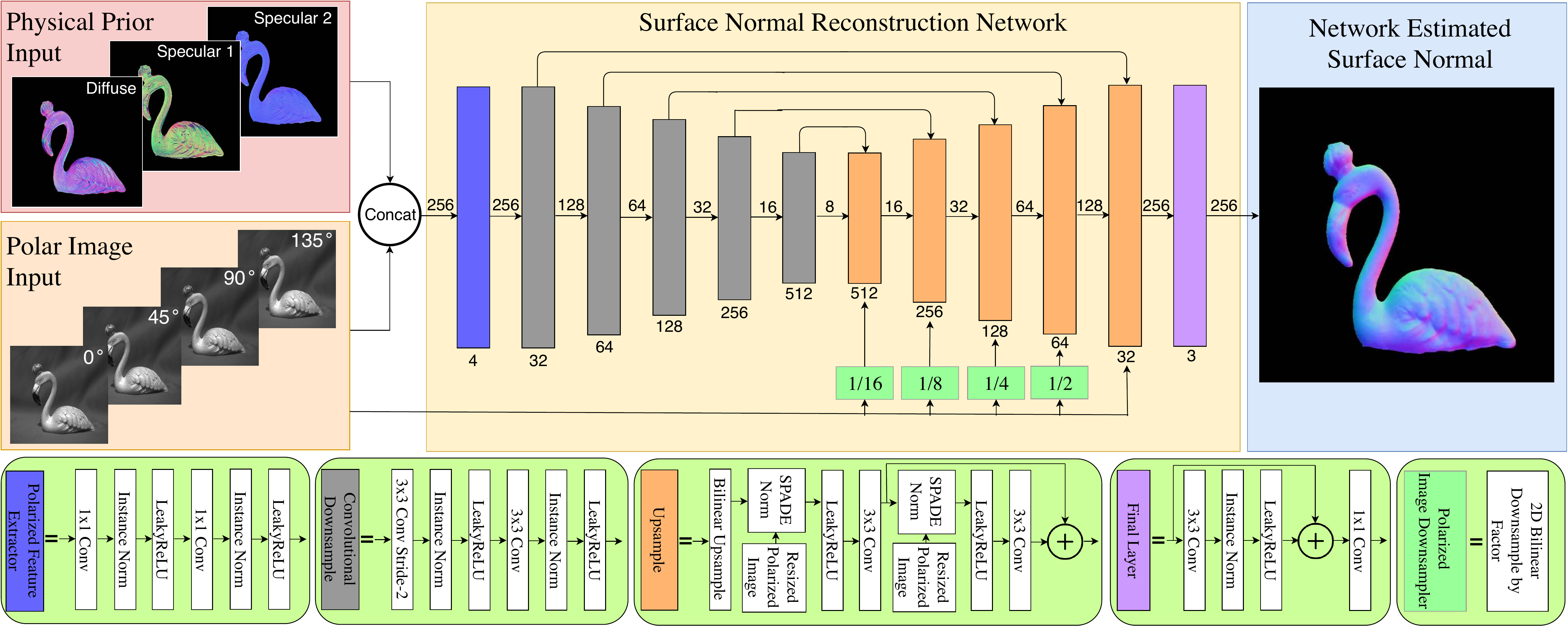}
    \caption{ \textbf{Overview of our proposed physics-based neural network.} The network is designed according to the encoder-decoder architecture in a fully convolutional manner. The blocks comprising the network are shown below the high-level diagram of our network pipeline. We use a block based on spatially-adaptive normalization as previously implemented in ~\cite{park2019semantic}. The numbers below the blocks refer to the number of output channels and the numbers next to the arrows refer to the spatial dimension.}
\label{fig:network structure}
\end{figure*}

A straightforward approach to estimating the normals, from polarization would be to simply take the set of polarization images as input, encode it into a feature map using a CNN, and feed the feature map into a normal-regression sub-network. Unsurprisingly, we find this results in normal reconstructions with higher MAE and undesirable lighting artifacts (see Figure~\ref{fig:priornoprior}). To guide the network towards more optimal solutions from the polarization information, one possible method is to force our learned solutions to adhere to the polarization equations described in Section~\ref{physical_solutions}, similar to the method used in ~\cite{Karpatne2017}. However, it is difficult to use these physical solutions for SfP tasks due to the following reasons: \textbf{1.} Normals derived from the equations will inherently have ambiguous azimuth angles. \textbf{2.} Specular reflection and diffuse reflection coexist simultaneously, and determining the proportion of each type is complicated. \textbf{3.} Polarization images are usually noisy, causing error in the ambiguous normals, especially when the degree of polarization is low. Shifting the azimuth angles by $\pi$ or $\frac{\pi}{2}$ could not reconstruct the surface normals properly for noisy images.

Therefore, we propose directly feeding both the polarization images and ambiguous normal maps into the network, and leave the network to learn how to combine both of these inputs effectively from training data. The estimated surface normals can be structured as following:
\begin{equation} 
  \boldsymbol{\hat{N}} =  f(\boldsymbol{I}_{\phi_{1}}, \boldsymbol{I}_{\phi_{2}}, ..., \boldsymbol{I}_{\phi_{M}}, \boldsymbol{N}_{diff}, \boldsymbol{N}_{spec1}, \boldsymbol{N}_{spec2}),
\end{equation}
where $f(\cdot)$ is the proposed prediction model, \{$\boldsymbol{I}_{\phi_{1}}$, $\boldsymbol{I}_{\phi_{2}}$, ..., $\boldsymbol{I}_{\phi_{M}}$\} is a set of polarization images, and $\boldsymbol{\hat{N}}$ is the estimated surface normals. We use the diffuse model in Section~\ref{physical_solutions} to calculate $\boldsymbol{N}_{diff}$, and $\boldsymbol{N}_{spec1}, \boldsymbol{N}_{spec2}$ are the two solutions from the specular model. These ambiguous normals can implicitly direct the proposed network to learn the surface normal information from the polarization.

Our network structure is illustrated in Figure~\ref{fig:network structure}. It consists of a fully-convolutional encoder to extract and combine high-level features from the ambiguous physical solutions and the polarization images, and a decoder to output the estimated normals, $\boldsymbol{\hat{N}}$. Although three polarization images are sufficient to capture the polarization information, we use images with a polarizer at $\phi_{pol} \in \{0\degree, 45\degree, 90\degree, 135\degree\}$. These images are concatenated channelwise with the ambiguous normal solutions as the model input.
   
Note that the fixed nature of our network input is not arbitrary, but based on the output of standard polarization cameras. Such cameras utilize a layer of polarizers above the photodiodes to capture these four polarization images in a single shot. Our network design is intended to enable applications using this  current single-shot capture technology. Single-shot capture is a clear advantage of the proposed method over alternative reconstruction approaches, such as photometric stereo, since it allows images to be captured in a less constrained manner.

\begin{figure}[t]
\centering
\includegraphics[width=\columnwidth]{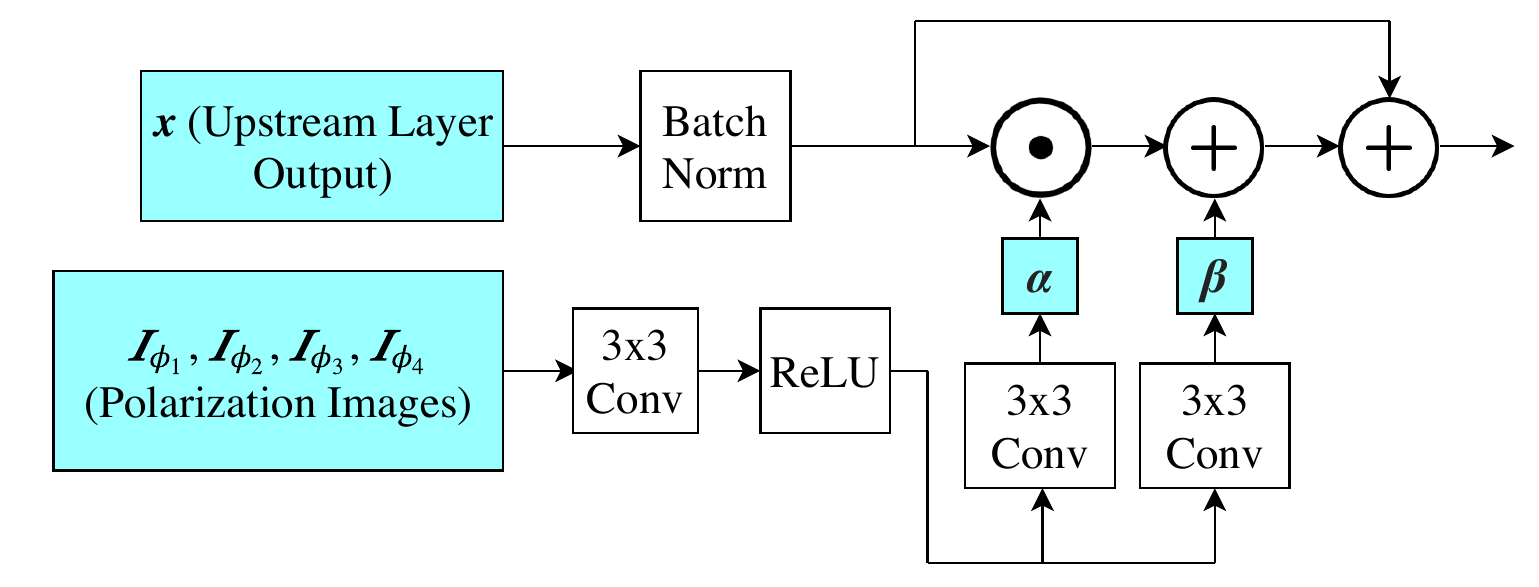}
    \caption{\textbf{Diagram of SPADE normalization block.} We use the resized polarization images to hierarchically inject back information in upsampling. The SPADE block, which takes a feature map $\boldsymbol{x}$ and a set of downsampled polarization images \{$\boldsymbol{{I}}_{\phi_{1}}$, $\boldsymbol{{I}}_{\phi_{2}}$, $\boldsymbol{{I}}_{\phi_{3}}$, $\boldsymbol{{I}}_{\phi_{4}}$\} as the input, learns affine modulation parameters $\boldsymbol{\alpha}$ and $\boldsymbol{\beta}$. The circle dot sign represents elementwise multiplication, and the circle plus sign represents elementwise addition.}
    \label{fig:SPADE}
\end{figure}

Before the downsampling layers of the encoder, we apply a series of $1 \times 1$ convolutional layers on the concatenated images and solutions, forcing the network to extract per-pixel polarization information. After polarization feature extraction, there are 5 encoder blocks to encode the input to a $B \times 512 \times 8 \times 8$ tensor, where $B$ is the minibatch size. The encoded tensor is then decoded by the same number of decoder blocks, with skip connections between blocks at the same hierarchical level as proposed in U-Net~\cite{ronneberger2015u}. It has been noted that such deep architectures may wash away some necessary information from the input~\cite{HuangStochastic, kumarHighwayNetworks}, so we apply spatially-adaptive normalization (SPADE)~\cite{park2019semantic} to address this problem. Motivated by their architecture, we replace the modulation parameters of batch normalization layers~\cite{ioffe2015batch} in each decoder block with parameters learned from downsampled polarization images using simple, two-layer convolutional sub-networks. The details of our adaptations to the SPADE module are depicted in Figure~\ref{fig:SPADE}.

Lastly, we normalize the output estimated normal vectors to unit length, and apply the cosine similarity loss function:
\begin{equation} 
  L_{cosine} =  \frac{1}{W \times H} \sum_{i}^{W}\sum_{j}^{H}(1 - \langle \boldsymbol{\hat{N}}_{ij}, \boldsymbol{N}_{ij} \rangle),
\end{equation}
where $\langle \cdot, \cdot \rangle$ denotes the dot product, $\boldsymbol{\hat{N}}_{ij}$ is the estimated surface normal at pixel location $(i, j)$, and $\boldsymbol{N}_{ij}$ is the corresponding ground truth surface normal. This loss is minimized when $\boldsymbol{\hat{N}}_{ij}$ and $\boldsymbol{N}_{ij}$ have identical orientation.

\section{Dataset and Implementation Details}

In what follows, we describe the dataset capture and organization as well as software implementation details. This is the first real-world dataset of its kind in the shape from polarization domain, containing polarization images and corresponding ground truth surface normals for a variety of objects, under multiple different lighting conditions. The Deep Shape from Polarization dataset can thus provide a baseline for future attempts at applying learning to the SfP problem.

\subsection{Dataset}\label{dataset}

\begin{figure*}[t]
  \includegraphics[width=\textwidth]{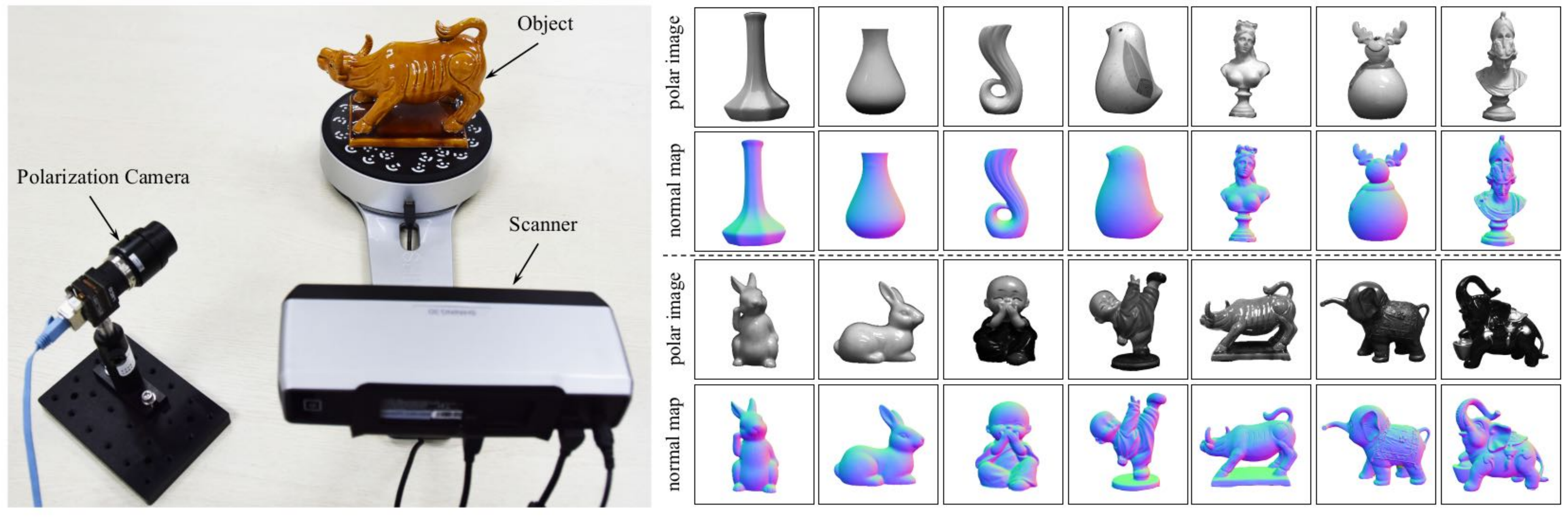}
    \caption{\textbf{This is the first dataset of its kind for the shape from polarization problem.} The capture setup and several example objects are shown above. We use a polarization camera to capture four gray-scale images of an object with four polarization angles in a single shot. The scanner is put next to the camera for obtaining the 3D shape of the object. The polarization images shown have a polarizer angle of $0$ degrees. The corresponding normal maps are aligned below. For each object, the capture process was repeated for 4 different orientations (front, back, left, right) and under 3 different lighting conditions (indoor lighting, outdoor overcast, and outdoor sunlight).}
\label{fig:setup_dataset}
\end{figure*}

A polarization camera~\cite{lucid} with a layer of polarizers above the photodiodes (as described in Section~\ref{learning_with_physics}) is used to capture four polarization images at angles $0\degree, 45\degree, 90\degree$ and $ 135\degree$ in a single shot. Then a structured light based 3D scanner~\cite{3DScanner} (with single shot accuracy no more than 0.1 mm, point distance from 0.17 mm to 0.2 mm, and a synchronized turntable for automatically registering scanning from multiple viewpoints) is used to obtain high-quality 3D shapes. Our real data capture setup is shown in Figure~\ref{fig:setup_dataset}. The scanned 3D shapes are aligned from the scanner's coordinate system to the image coordinate system of the polarization camera by using the shape-to-image alignment method adopted in~\cite{Shi19}. Finally, we compute the surface normals of the aligned shapes by using the Mitsuba renderer~\cite{Mitsuba}.  Our introduced dataset consists of 25 different objects, each object with 4 different orientations for a total of 100 object-orientation combinations. For each  object-orientation combination, we capture images in 3 lighting conditions: indoors, outdoors on an overcast day, and outdoors on a sunny day. In total, we capture 300 images for this dataset, each with 4 polarization angles. 

\subsection{Software implementation}

Our model was implemented in PyTorch~\cite{paszke2017automatic}, and trained for 500 epochs with a batch size of 4. It took around 8 hours for the network to converge with a single NVIDIA GeForce RTX 2070. We used the Adam optimizer~\cite{kingma2014adam} with default parameters with a base learning rate of 0.01. Considering the image size, we trained our model on image patches, which is relatively common in shape estimation tasks~\cite{xiong2014shading, mo2018uncalibrated}. During training, we randomly crop images to $256 \times 256$ patches for the data augmentation purpose. During testing, we split each image into $256 \times 256$ patches and rejoin the estimated normal patches from our network output. The final prediction is the average of 32 cropped predictions with shifted input to preserve the accuracy at patch boundaries. 

\begin{figure}[t]
    \centering
    \includegraphics[width=\columnwidth]{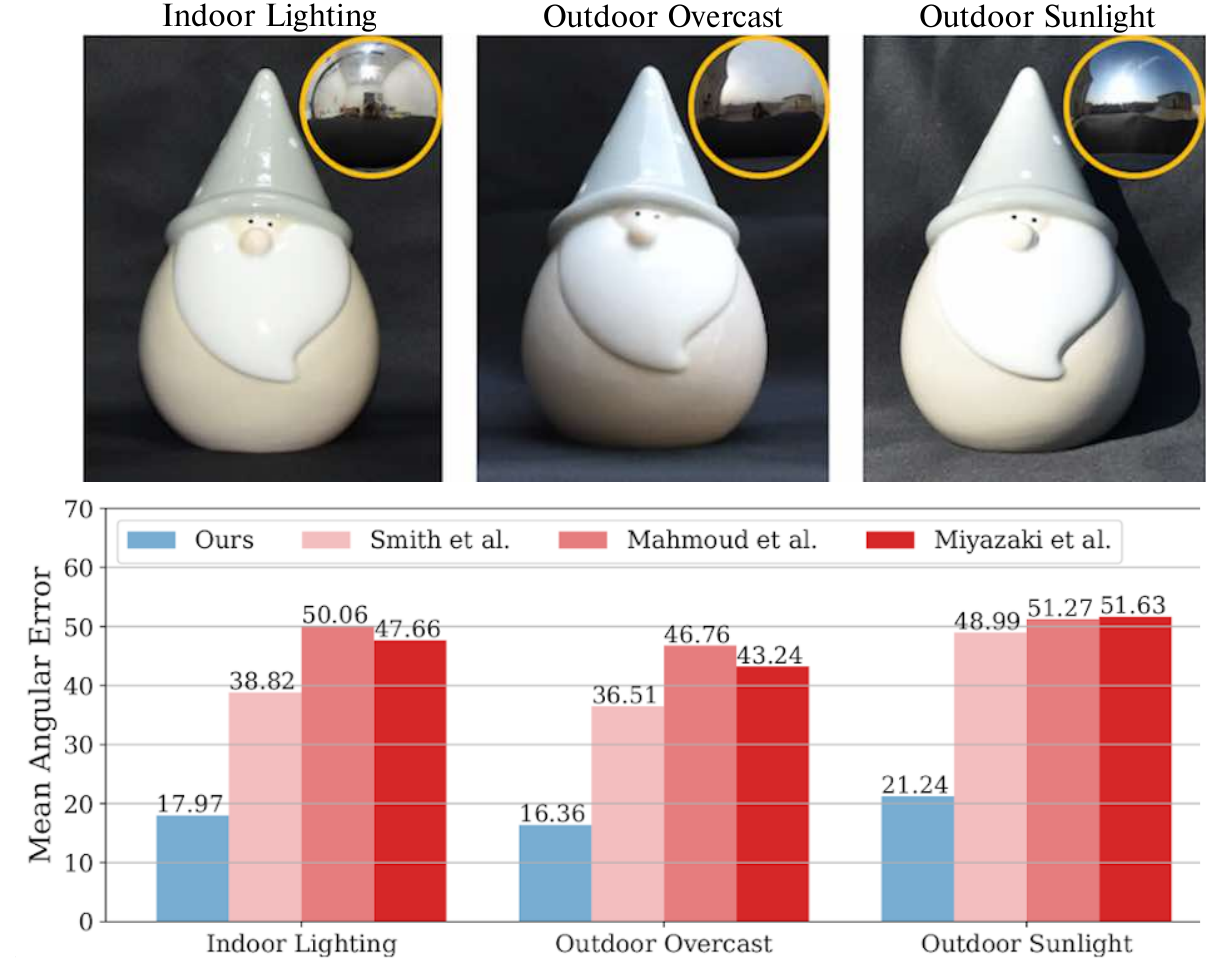}
    \caption{\textbf{The proposed method handles objects under varied lighting conditions}. Note that our method has very similar mean angular error among all test objects across the three lighting conditions (bottom row).} 
    \label{fig:illumination}
\end{figure}

\section{Experimental Results}

In this section, we evaluate our model with the presented challenging real-world scene benchmark, and compare it against three physics-only methods for SfP. All neural networks were trained on the same training data as discussed in Section~\ref{dataset}. We additionally train on the physics-based priors for the best-performing method we propose. To quantify shape accuracy, we compute the widely used mean angular error (MAE) score on the surface normals. 

\subsection{Comparisons to physics-based SfP} We used a test dataset consisting of scenes that include \textsc{ball}, \textsc{horse}, \textsc{vase}, \textsc{christmas}, \textsc{flamingo}, \textsc{dragon}. On this test set, we implement three physics-based methods for SfP as a baseline: \textbf{1.} Smith \textit{et al.}~\cite{Smith18}. \textbf{2.} Mahmoud \textit{et al.}~\cite{mahmoud2012direct}. \textbf{3.} Miyazaki \textit{et al.}~\cite{miyazaki2003polarization}. The first method recovers the depth map directly, and we only use the diffuse model due to the lack of specular reflection masks. The surface normals are obtained from the estimated depth with bicubic fit. Both the first and the second methods require lighting input, and we use the estimated lighting from the first method during comparison. The second method also requires known albedo, and following convention, we assume a uniform albedo of 1. Note the method proposed in ~\cite{miyazaki2003polarization} is the same as that presented in ~\cite{atkinson2006recovery}. We omit comparison with Tozza \textit{et al.}~\cite{Tozza17}, as it requires two unpolarized intensity images, with two different light source directions. To motivate a fair comparison, we obtained the comparison codes directly from Smith \textit{et al.}~\cite{Smith18}. \footnote{\url{https://github.com/waps101/depth-from-polarisation}}

\subsection{Robustness to lighting variations} Figure~\ref{fig:illumination} shows the robustness of the method to various lighting conditions. Our dataset includes lighting in three broad categories: (a) indoor lighting; (b) outdoor overcast; and (c) outdoor sunlight. The proposed method has the lowest MAE, over the three lighting conditions. Furthermore, our method is consistent across conditions, with only slight differences in MAE for each object between lightings.

\begin{figure} [t]
    \centering
    \includegraphics[width=\columnwidth]{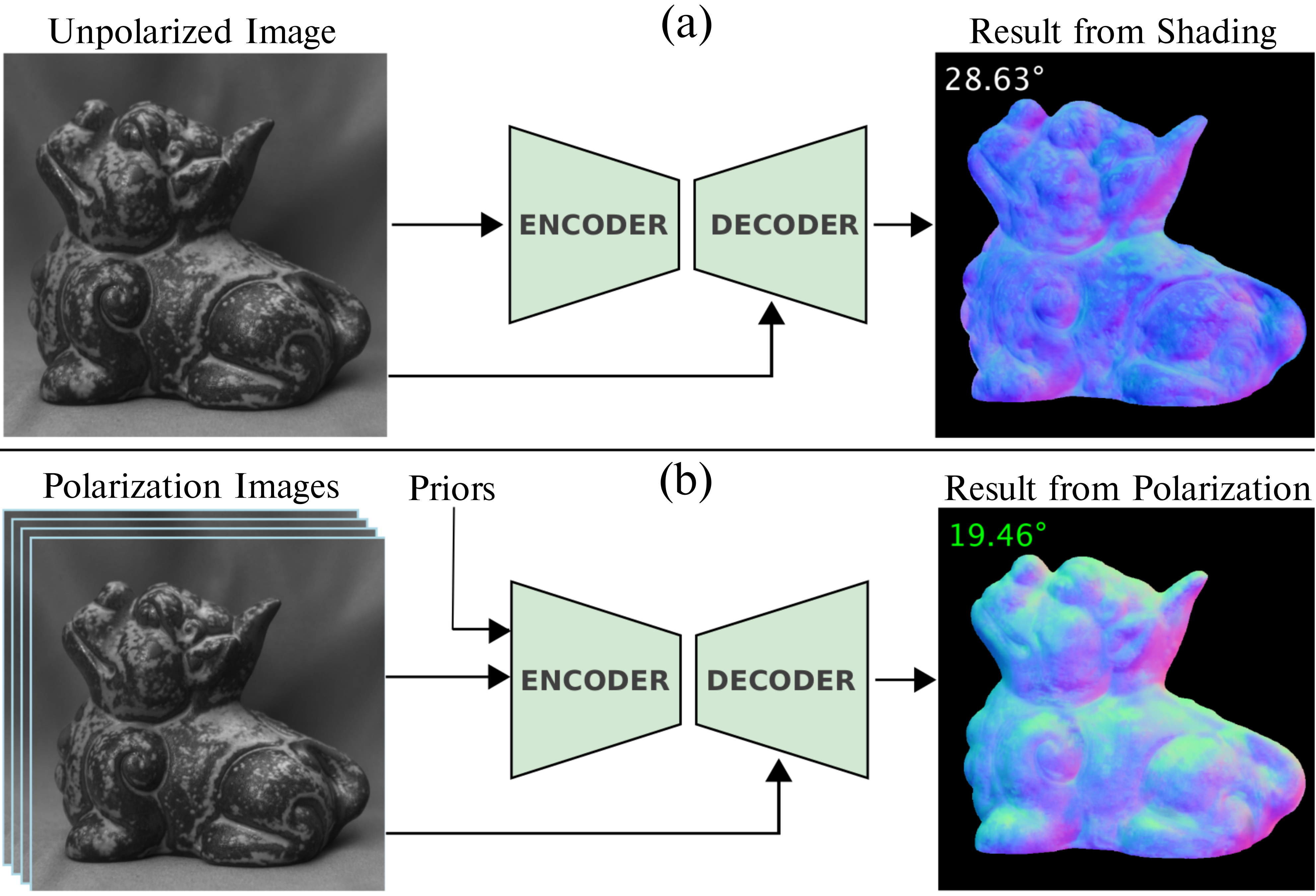}
    \caption {\textbf{Our network is learning from polarization cues, not just shading cues.} An ablation study conducted on the \textsc{Dragon} scene. In (a) the network does not have access to polarization inputs. In (b) the network can learn from polarization inputs and polarization physics. Please refer to Figure~\ref{fig:bigresult}, row c, for the ground truth shape of the \textsc{Dragon}.
     } 
    \label{fig:PolVsShade}
\end{figure}

\subsection{Importance of polarization} An interesting question is how much of the shape information is learned from polarization cues as compared to shading cues. Figure~\ref{fig:PolVsShade} explores the benefit of polarization by ablating the network inputs. We compare two cases. Figure~\ref{fig:PolVsShade}(a) shows the resulting shape reconstruction when using a network architecture optimized for an unpolarized image input. The shape has texture copy and a high MAE of 28.63 degrees. In contrast, Figure~\ref{fig:PolVsShade}(b) shows shape reconstruction from our proposed method of learning from four polarization images and a model of polarization physics. We observe that shape reconstruction using polarization cues is more robust to texture copy artifacts, and has a lower MAE of only 19.46 degrees. Although only one image is used in the shading network (as is typical for shape from shading), this image is computed using an average of the four polarization images. Thus the distinction between the two cases in Figure~\ref{fig:PolVsShade}(a) and~\ref{fig:PolVsShade}(b)  is the polarization diversity, rather than improvements in photon noise.

\begin{figure}[t]
    \centering
    \includegraphics[width=\columnwidth]{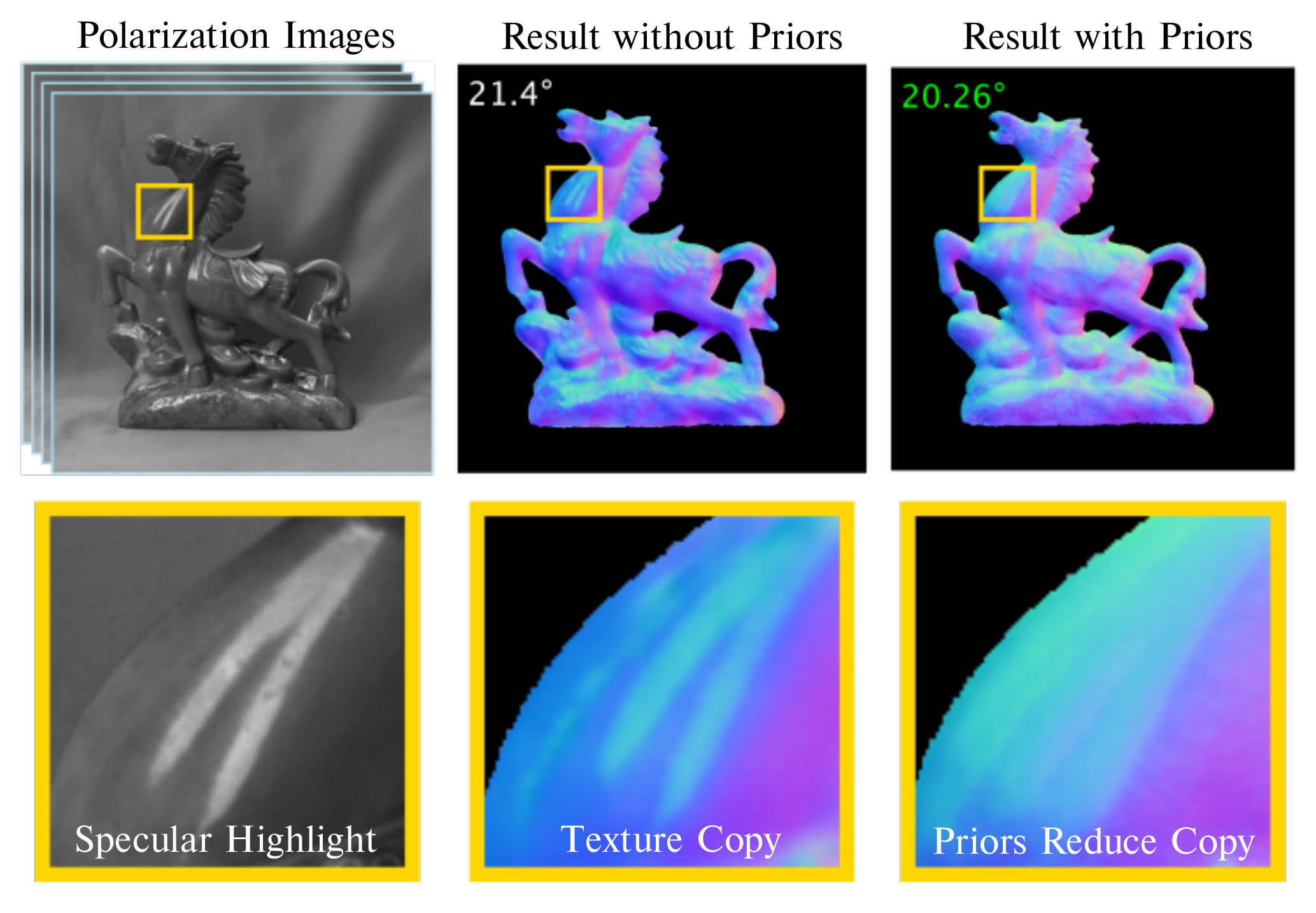}
    \caption {\textbf{Ablation test shows that the physics-based prior reduces texture copy artifacts.}  We see that the specular highlight in the input polarization image is directly copied into the normal reconstruction without priors. Note that our prior-based method shows stronger suppression of the copy artifact.  } 
    \label{fig:priornoprior}
\end{figure}

\subsection{Importance of physics revealed by ablating priors} Figure~\ref{fig:priornoprior} highlights the importance of physics-based learning, as compared to traditional machine learning. Here, we refer to ``traditional machine learning'' as learning shape using only the polarization images as input. These results are shown in the middle column of Figure~\ref{fig:priornoprior}. Shape reconstructions based on traditional machine learning exhibit image-based artifacts, because the polarization images contain brightness variations that are not due to geometry, but due to specular highlights (e.g., the \textsc{horse} is shiny). Learning from just the polarization images alone causes these image-based variations to masquerade as shape variations, as shown in the zoomed inset of  Figure~\ref{fig:priornoprior}. A term used for this is \emph{texture copy}, where image texture is undesirably copied onto the geometry~\cite{kadambi2015polarized}. In contrast, the proposed results with physics priors are shown in the rightmost inset of Figure~\ref{fig:priornoprior}, showing less dependence on image-based texture (because we also input the geometry-based physics model). 

\subsection{Quantitative evaluation on our test set}\label{section:quant}

We use MAE\footnote{MAE is the most commonly reported measure for surface normal reconstruction. However, in many cases it is a deceptive metric. We find that a few outliers in high-frequency regions can skew the MAE for entire reconstructions. Accordingly, we emphasize the qualitative comparisons of the proposed method to its physics-based counterparts.} to make a quantitative comparison between our method and the previous physics-based approaches. Table~\ref{tab:numbers} shows that the proposed method has the lowest MAE on each object, as well as the overall test set. The two most challenging scenes in the test set are the \textsc{horse} and the \textsc{dragon}. The former has intricate detail and specularities, while the latter has a mixed material surface. The physics-based methods struggle on these challenging scenes as all scenes have over 49 degrees of mean angular error. The method from Smith \textit{et al.}~\cite{Smith18} has the second-lowest error on the \textsc{dragon} scene, but the method from Miyazaki \textit{et al.}~\cite{miyazaki2003polarization} has the second-lowest error on the \textsc{horse} scene. On the overall test set, the physics-based methods are all clustered between 41.4 and 49.0 degrees, while the physics-based deep learning approach we propose achieves over a two-fold reduction in error to 18.5 degrees.

It is worth noting is that the many of the physics-based methods use models which may not hold well for a varied, real world scenes. The result from Smith \textit{et al.}~\cite{Smith18} assumes a simple hybrid reflection model and uses a least squares combinatorial lighting estimation. This physical assumption may not be well satisfied and the lighting estimation may be inaccurate for many unconstrained, real world environments, and the estimated depth becomes inaccurate, which results in a normal map with a larger error. The method of Mahmoud \textit{et al.}~\cite{mahmoud2012direct} uses shading constraints that assume a distant light source, which is not the case for some of the tested scenes, especially the indoor ones.  Finally, the large region-wise anomalies on many of the results from Miyazaki \textit{et al.}~\cite{miyazaki2003polarization} are due to the region-growing constraint on convexity that is imposed.

\begin{table}
\caption{
\textbf{Our method outperforms previous methods for each object in the test set.} Numbers represent the MAE averaged across the three lighting conditions for each object. The best model is marked in {\color{magenta}magenta} and the second-best is in {\color{blue}blue}.}

\centering
\resizebox{.48\textwidth}{!}{%

\begin{tabular}{lcccc}
\hline

Scene & \textbf{Proposed} & Smith~\cite{Smith18} & Mahmoud~\cite{mahmoud2012direct} & Miyazaki~\cite{miyazaki2003polarization} \\ \hline
 \textsc{Box} & {\textbf{\color{magenta}23.31\degree}} & {\textbf{\color{blue} 31.00\degree}} & 41.51\degree & 45.47\degree \\
 \textsc{Dragon} & {\textbf{\color{magenta}21.55\degree}} & {\textbf{\color{blue} 49.16\degree}} & 70.72\degree & 57.72\degree \\

 \textsc{Father Christmas} & {\textbf{\color{magenta}13.50\degree}} & 39.68\degree& {\textbf{\color{blue} 39.20\degree}}  & 41.50\degree \\
 \textsc{Flamingo} & {\textbf{\color{magenta}20.19\degree}} & {\textbf{\color{blue} 36.05\degree}} & 47.98\degree & 45.58\degree \\
 \textsc{Horse} & {\textbf{\color{magenta}22.27\degree}} & 55.87\degree & {\textbf{\color{blue}50.55}}\degree &  51.34\degree \\
 \textsc{Vase} & {\textbf{\color{magenta}10.32\degree}} & {\textbf{\color{blue} 36.88\degree}} & 44.23\degree & 43.47\degree  \\
 \textsc{Whole Set} & {\textbf{\color{magenta}18.52\degree}} & {\textbf{\color{blue} 41.44\degree}} & 49.03\degree & 47.51\degree  \\

 \hline
\end{tabular}
}
\label{tab:numbers}
\end{table}

\begin{figure*}[h]
\includegraphics[width=1.02\textwidth]{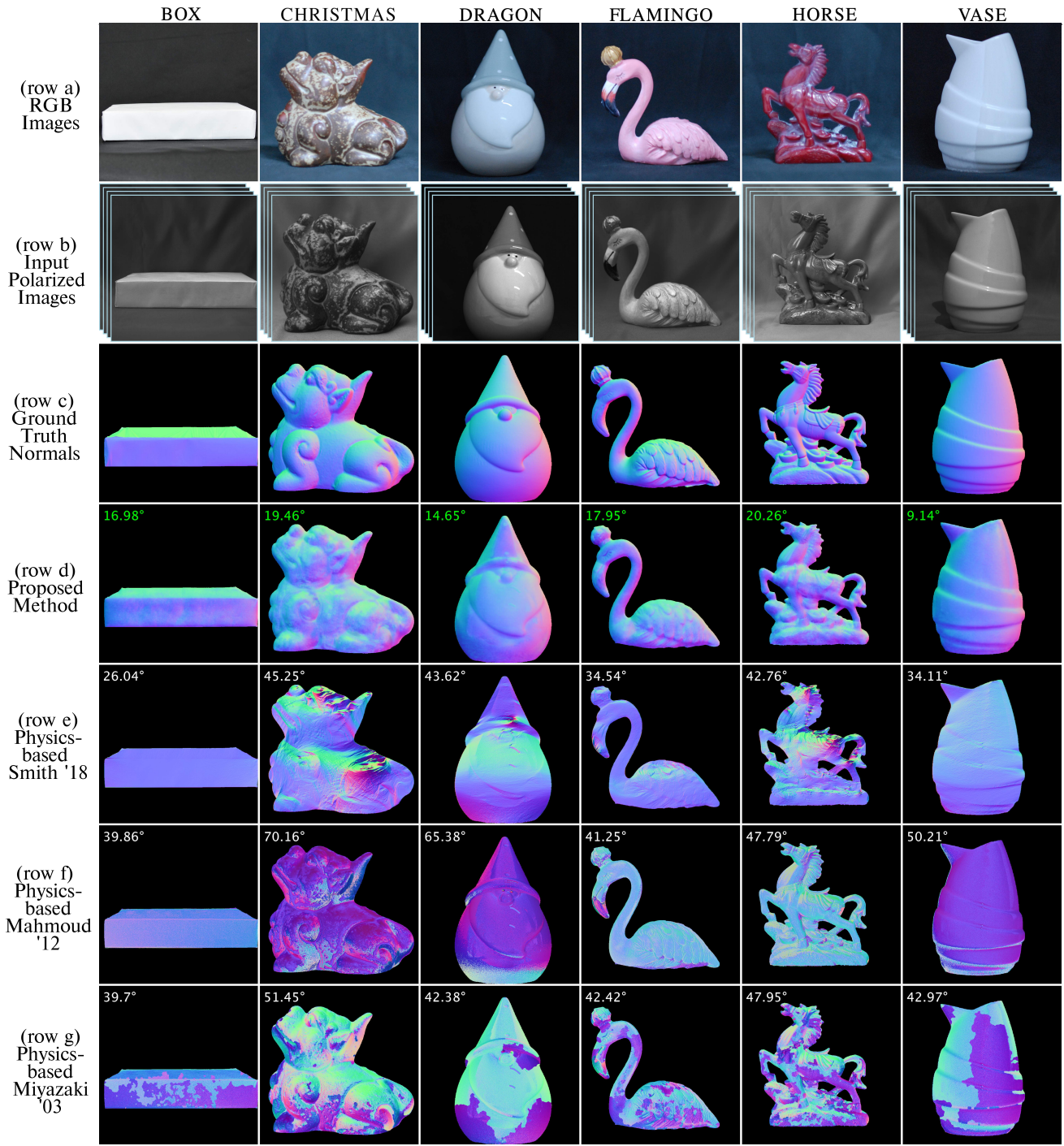}
\caption{\textbf{The proposed method shows qualitative and quantitative improvements in shape recovery on various objects from our test dataset.} (row a) Shows the RGB scene photographs for context - these are not used as the input to any of the methods. (row b) The input to all methods are a stack of four polarization photographs at angles of 0\degree, 45\degree, 90\degree, and 135\degree (row c). The ground truth normals, obtained experimentally. (row d) The proposed approach for shape recovery. (row e-g) We compare with physics-based SfP methods by Smith \textit{et al.}~\cite{smith2016linear}, Mahmoud \textit{et al.}~\cite{mahmoud2012direct} and Miyazaki \textit{et al.}~\cite{miyazaki2003polarization}. (We omit the results from Atkinson \textit{et al.}~\cite{atkinson2006recovery}, which uses a similar method as~\cite{miyazaki2003polarization}). The MAE for the reconstructions is shown on the top left of each cell in (row c) - (row g).}
\label{fig:bigresult}
\end{figure*}

\subsection{Qualitative evaluation on our test set} \label{section:qual} Figure~\ref{fig:bigresult} shows qualitative and quantitative data for various objects in our test set. The RGB images in (row a) are not used as input, but are shown in the top row of the figure for context about material properties. The input to all the methods shown is four polarization images, shown in (row b) of Figure~\ref{fig:bigresult}. The ground truth shape is shown in (row c), and corresponding shape reconstructions for the proposed method are shown in (row d). Comparison methods are shown in (row e) through (row g). It is worth noting that the physics-based methods particularly struggle with \emph{texture copy} artifacts, where color variations masquerade as geometric variations. This can be seen in Figure~\ref{fig:bigresult}, (row f), where the physics-based reconstruction of Mahmoud~\cite{mahmoud2012direct} confuses the color variation in the beak of the \textsc{flamingo} with a geometric variation. In contrast, our proposed method, shown in (row d), recovers the beak more accurately. Beyond texture copy, another limitation of physics-based methods lies in the difficulty of solving the \emph{ambiguity problem}, discussed earlier in this paper. In row g, the physics-based approach from Miyazaki \textit{et al.}~\cite{miyazaki2003polarization} has significant ambiguity errors. This can be seen as the fixed variations in color of normal maps, which are not due to random noise. Although less drastic, the physics-based method of Smith \textit{et al.}~\cite{Smith18} also shows such fixed pattern artifacts, due to the underdetermined nature of the problem. Our proposed method is fairly robust to fixed pattern error, and our deviation from ground truth is largely in areas with high-frequency detail. Although the focus of Figure~\ref{fig:bigresult} is to highlight qualitative comparisons, it is worth noting that the MAE in of the proposed method is the lowest for all these scenes (lowest MAE is highlighted in green font).

\section{Discussion}

In summary, we presented a first attempt re-examining Shape from Polarization through the lens of deep learning, and specifically, physics-based deep learning. Table~\ref{tab:numbers} shows that our network achieves over two-fold reduction in shape error, from 41.4 degrees~\cite{Smith18} to 18.5 degrees. An ablation test verifies the importance of using the physics-based prior in the deep learning model. We also observe that the proposed model performs well under varied lighting conditions, while previous physics based approaches have higher error and more variation. \newline
\newline
\noindent\textbf{Future Work} The framerate of our technique is limited both by the feed-forward pass, as well as the time required to calculate the physical prior (about 1 second per frame). Future work could explore parallelizing the physics-based calculations or using approximations for more efficient compute. The lessons learned in this ``Deep Shape from Polarization'' study may also apply to a future ``Deep Polarized 3D'' study, which has previously been performed with a physics-based matrix inversion~\cite{kadambi2015polarized}. As discussed in Section~\ref{section:quant}, the high MAE is largely due to a few regions with extremely fine detail. Finding ways to effectively weight these areas more heavily or add a refinement stage focused on these challenging regions, are promising avenues for future exploration. Finally, while this is an important first dataset for SfP, we look forward to adding addition lighting conditions and objects with more inter-reflections. 
\newline
\newline
\noindent\textbf{Conclusion} We hope the results of this study encourage future explorations at the seamline of deep learning and polarization. We believe our work sheds light on how deep architectures can learn to overcome ambiguities in known physical constraints, but leaves room for further improvement upon utilization of data with ill-posed physics. 

\clearpage

{\small
\bibliographystyle{ieee_fullname}
\bibliography{egbib}

\begin{thebibliography}{10}\itemsep=-1pt

\bibitem{Atkinson17}
Gary~A. Atkinson.
\newblock Polarisation photometric stereo.
\newblock {\em Computer Vision and Image Understanding}, 2017.

\bibitem{Atkinson18}
Gary~A. Atkinson and J\"{u}rgen~D. Ernst.
\newblock High-sensitivity analysis of polarization by surface reflection.
\newblock {\em Machine Vision and Applications}, 2018.

\bibitem{Atkinson05}
Gary~A. Atkinson and Edwin~R. Hancock.
\newblock Multi-view surface reconstruction using polarization.
\newblock {\em ICCV}, 2005.

\bibitem{atkinson2006recovery}
Gary~A Atkinson and Edwin~R Hancock.
\newblock Recovery of surface orientation from diffuse polarization.
\newblock {\em IEEE TIP}, 2006.

\bibitem{baek2018simultaneous}
Seung-Hwan Baek, Daniel~S Jeon, Xin Tong, and Min~H Kim.
\newblock Simultaneous acquisition of polarimetric {SVBRDF} and normals.
\newblock {\em ACM SIGGRAPH (TOG)}, 2018.

\bibitem{Berger17}
Kai Berger, Randolph Voorhies, and Larry~H. Matthies.
\newblock Depth from stereo polarization in specular scenes for urban robotics.
\newblock {\em ICRA}, 2017.

\bibitem{Chen18PS}
Guanying Chen, Kai Han, and Kwan-Yee~K. Wong.
\newblock {PS-FCN}: A flexible learning framework for photometric stereo.
\newblock {\em ECCV}, 2018.

\bibitem{Chen18}
Lixiong Chen, Yinqiang Zheng, Art Subpa-asa, and Imari Sato.
\newblock Polarimetric three-view geometry.
\newblock {\em ECCV}, 2018.

\bibitem{Cui17}
Zhaopeng Cui, Jinwu Gu, Boxin Shi, Ping Tan, and Jan Kautz.
\newblock Polarimetric multi-view stereo.
\newblock {\em CVPR}, 2017.

\bibitem{deschaintre2018single}
Valentin Deschaintre, Miika Aittala, Fredo Durand, George Drettakis, and Adrien
  Bousseau.
\newblock Single-image {SVBRDF} capture with a rendering-aware deep network.
\newblock {\em ACM SIGGRAPH (TOG)}, 2018.

\bibitem{Drbohlav01}
Ondrej Drbohlav and Radim Sara.
\newblock Unambiguous determination of shape from photometric stereo with
  unknown light sources.
\newblock {\em ICCV}, 2001.

\bibitem{ghosh2010circularly}
Abhijeet Ghosh, Tongbo Chen, Pieter Peers, Cyrus~A Wilson, and Paul Debevec.
\newblock Circularly polarized spherical illumination reflectometry.
\newblock {\em ACM SIGGRAPH (TOG)}, 2010.

\bibitem{ghosh2011multiview}
Abhijeet Ghosh, Graham Fyffe, Borom Tunwattanapong, Jay Busch, Xueming Yu, and
  Paul Debevec.
\newblock Multiview face capture using polarized spherical gradient
  illumination.
\newblock {\em ACM SIGGRAPH (TOG)}, 2011.

\bibitem{guarnera2012estimating}
Giuseppe~Claudio Guarnera, Pieter Peers, Paul Debevec, and Abhijeet Ghosh.
\newblock Estimating surface normals from spherical stokes reflectance fields.
\newblock {\em ECCV}, 2012.

\bibitem{HuangStochastic}
Gao Huang, Yu Sun, Zhuang Liu, Daniel Sedra, and Kilian Weinberger.
\newblock Deep networks with stochastic depth.
\newblock {\em CoRR}, 2016.

\bibitem{Huynh10}
Cong~Phuoc Huynh, A. Robles-Kelly, and Edwin~R. Hancock.
\newblock Shape and refractive index recovery from single-view polarisation
  images.
\newblock {\em CVPR}, 2010.

\bibitem{Huynh13}
Cong~Phuoc Huynh, A. Robles-Kelly, and Edwin~R. Hancock.
\newblock Shape and refractive index from single-view spectro-polarimetric
  images.
\newblock {\em IJCV}, 2013.

\bibitem{Ikehata18}
Satoshi Ikehata.
\newblock {CNN-PS}: {CNN}-based photometric stereo for general non-convex
  surfaces.
\newblock {\em ECCV}, 2018.

\bibitem{ioffe2015batch}
Sergey Ioffe and Christian Szegedy.
\newblock Batch normalization: Accelerating deep network training by reducing
  internal covariate shift.
\newblock {\em arXiv preprint arXiv:1502.03167}, 2015.

\bibitem{Mitsuba}
Wenzel Jakob.
\newblock Mitsuba renderer, 2010.
\newblock http://www.mitsuba-renderer.org.

\bibitem{kadambi2015polarized}
Achuta Kadambi, Vage Taamazyan, Boxin Shi, and Ramesh Raskar.
\newblock Polarized 3{D}: High-quality depth sensing with polarization cues.
\newblock {\em ICCV}, 2015.

\bibitem{Kadambi17}
Achuta Kadambi, Vage Taamazyan, Boxin Shi, and Ramesh Raskar.
\newblock Depth sensing using geometrically constrained polarization normals.
\newblock {\em IJCV}, 2017.

\bibitem{Karpatne2017}
Anuj Karpatne, William Watkins, Jordan Read, and Vipin Kumar.
\newblock Physics-guided neural networks {(PGNN):} an application in lake
  temperature modeling.
\newblock {\em CoRR}, 2017.

\bibitem{kingma2014adam}
Diederik~P Kingma and Jimmy Ba.
\newblock Adam: A method for stochastic optimization.
\newblock {\em arXiv preprint arXiv:1412.6980}, 2014.

\bibitem{li2017modeling}
Xiao Li, Yue Dong, Pieter Peers, and Xin Tong.
\newblock Modeling surface appearance from a single photograph using
  self-augmented convolutional neural networks.
\newblock {\em ACM SIGGRAPH (TOG)}, 2017.

\bibitem{li2018materials}
Zhengqin Li, Kalyan Sunkavalli, and Manmohan Chandraker.
\newblock Materials for masses: {SVBRDF} acquisition with a single mobile phone
  image.
\newblock {\em ECCV}, 2018.

\bibitem{li2018learning_sv}
Zhengqin Li, Zexiang Xu, Ravi Ramamoorthi, Kalyan Sunkavalli, and Manmohan
  Chandraker.
\newblock Learning to reconstruct shape and spatially-varying reflectance from
  a single image.
\newblock {\em ACM SIGGRAPH Asia (TOG)}, 2018.

\bibitem{Lindell18}
David~B. Lindell, Matthew O'Toole, and Gordon Wetzstein.
\newblock Single-photon 3{D} imaging with deep sensor fusion.
\newblock {\em ACM SIGGRAPH (TOG)}, 2018.

\bibitem{lucid}
{Lucid Vision Phoenix polarization camera}.
\newblock \url{https://thinklucid.com/product/phoenix-5-0-mp-polarized-model/}.
\newblock 2018.

\bibitem{ma2007rapid}
Wan-Chun Ma, Tim Hawkins, Pieter Peers, Charles-Felix Chabert, Malte Weiss, and
  Paul Debevec.
\newblock Rapid acquisition of specular and diffuse normal maps from polarized
  spherical gradient illumination.
\newblock {\em Eurographics Conference on Rendering Techniques}, 2007.

\bibitem{maeda2018dynamic}
Tomohiro Maeda, Achuta Kadambi, Yoav~Y Schechner, and Ramesh Raskar.
\newblock Dynamic heterodyne interferometry.
\newblock {\em ICCP}, 2018.

\bibitem{mahmoud2012direct}
Ali~H Mahmoud, Moumen~T El-Melegy, and Aly~A Farag.
\newblock Direct method for shape recovery from polarization and shading.
\newblock {\em ICIP}, 2012.

\bibitem{marco2017deeptof}
Julio Marco, Quercus Hernandez, Adolfo Munoz, Yue Dong, Adrian Jarabo, Min~H
  Kim, Xin Tong, and Diego Gutierrez.
\newblock Deeptof: off-the-shelf real-time correction of multipath interference
  in time-of-flight imaging.
\newblock {\em ACM SIGGRAPH (TOG)}, 2017.

\bibitem{Miyazaki04}
Daisuke Miyazaki, Masataka Kagesawa, and Katsushi Ikeuchi.
\newblock Transparent surface modeling from a pair of polarization images.
\newblock {\em PAMI}, 2004.

\bibitem{Miyazaki16}
Daisuke Miyazaki, Takuya Shigetomi, Masashi Baba, Ryo Furukawa, Shinsaku Hiura,
  and Naoki Asada.
\newblock Surface normal estimation of black specular objects from multiview
  polarization images.
\newblock {\em International Society for Optics and Photonics, Optical
  Engineering}, 2016.

\bibitem{miyazaki2003polarization}
Daisuke Miyazaki, Robby~T Tan, Kenji Hara, and Katsushi Ikeuchi.
\newblock Polarization-based inverse rendering from a single view.
\newblock {\em ICCV}, 2003.

\bibitem{mo2018uncalibrated}
Zhipeng Mo, Boxin Shi, Feng Lu, Sai-Kit Yeung, and Yasuyuki Matsushita.
\newblock Uncalibrated photometric stereo under natural illumination.
\newblock In {\em Proceedings of the IEEE Conference on Computer Vision and
  Pattern Recognition}, pages 2936--2945, 2018.

\bibitem{Ngo15}
Trung~Thanh Ngo, Hajime Nagahara, and Rin{-}ichiro Taniguchi.
\newblock Shape and light directions from shading and polarization.
\newblock {\em CVPR}, 2015.

\bibitem{park2019semantic}
Taesung Park, Ming-Yu Liu, Ting-Chun Wang, and Jun-Yan Zhu.
\newblock Semantic image synthesis with spatially-adaptive normalization.
\newblock {\em CVPR}, 2019.

\bibitem{paszke2017automatic}
Adam Paszke, Sam Gross, Soumith Chintala, Gregory Chanan, Edward Yang, Zachary
  DeVito, Zeming Lin, Alban Desmaison, Luca Antiga, and Adam Lerer.
\newblock Automatic differentiation in pytorch.
\newblock {\em NIPS-W}, 2017.

\bibitem{PolarM}
{PolarM polarization camera}.
\newblock \url{http://www.4dtechnology.com/products/polarimeters/polarcam/}.
\newblock 2017.

\bibitem{riviere2017polarization}
J{\'e}r{\'e}my Riviere, Ilya Reshetouski, Luka Filipi, and Abhijeet Ghosh.
\newblock Polarization imaging reflectometry in the wild.
\newblock {\em ACM SIGGRAPH (TOG)}, 2017.

\bibitem{ronneberger2015u}
Olaf Ronneberger, Philipp Fischer, and Thomas Brox.
\newblock U-net: Convolutional networks for biomedical image segmentation.
\newblock {\em MICCAI}, 2015.

\bibitem{Santo17}
Hiroaki Santo, Masaki Samejima, Yusuke Sugano, Boxin Shi, and Yasuyuki
  Matsushita.
\newblock Deep photometric stereo network.
\newblock {\em ICCV Workshops}, 2017.

\bibitem{satat2017object}
Guy Satat, Matthew Tancik, Otkrist Gupta, Barmak Heshmat, and Ramesh Raskar.
\newblock Object classification through scattering media with deep learning on
  time resolved measurement.
\newblock {\em OSA Optics Express}, 2017.

\bibitem{schechner2015self}
Yoav~Y Schechner.
\newblock Self-calibrating imaging polarimetry.
\newblock {\em ICCP}, 2015.

\bibitem{sengupta2018sfsnet}
Soumyadip Sengupta, Angjoo Kanazawa, Carlos~D Castillo, and David~W Jacobs.
\newblock Sf{S}net: Learning shape, reflectance and illuminance of faces in the
  wild.
\newblock {\em CVPR}, 2018.

\bibitem{Shi19}
Boxin Shi, Zhipeng Mo, Zhe Wu, Dinglong Duan, Sai-Kit Yeung, and Ping Tan.
\newblock A benchmark dataset and evaluation for non-{L}ambertian and
  uncalibrated photometric stereo.
\newblock {\em PAMI}, 2019.

\bibitem{3DScanner}
{SHINING 3D scanner}.
\newblock \url{https://www.einscan.com/einscan-se-sp}.
\newblock 2018.

\bibitem{smith2016linear}
William A.~P. Smith, Ravi Ramamoorthi, and Silvia Tozza.
\newblock Linear depth estimation from an uncalibrated, monocular polarisation
  image.
\newblock {\em ECCV}, 2016.

\bibitem{Smith18}
William A.~P. Smith, Ravi Ramamoorthi, and Silvia Tozza.
\newblock Height-from-polarisation with unknown lighting or albedo.
\newblock {\em PAMI}, 2018.

\bibitem{kumarHighwayNetworks}
Rupesh~Kumar Srivastava, Klaus Greff, and J{\"{u}}rgen Schmidhuber.
\newblock Highway networks.
\newblock {\em CoRR}, 2015.

\bibitem{Su18}
Shuochen Su, Felix Heide, Gordon Wetzstein, and Wolfgang Heidrich.
\newblock Deep end-to-end {time-of-flight} imaging.
\newblock {\em CVPR}, 2018.

\bibitem{tancik2018flash}
Matthew Tancik, Guy Satat, and Ramesh Raskar.
\newblock Flash photography for data-driven hidden scene recovery.
\newblock {\em arXiv preprint arXiv:1810.11710}, 2018.

\bibitem{tancik2018data}
Matthew Tancik, Tristan Swedish, Guy Satat, and Ramesh Raskar.
\newblock Data-driven non-line-of-sight imaging with a traditional camera.
\newblock {\em OSA Imaging and Applied Optics}, 2018.

\bibitem{Taniai18}
Tatsunori Taniai and Takanori Maehara.
\newblock Neural inverse rendering for general reflectance photometric stereo.
\newblock {\em ICML}, 2018.

\bibitem{Teo_2018_CVPR}
Daniel Teo, Boxin Shi, Yinqiang Zheng, and Sai-Kit Yeung.
\newblock Self-calibrating polarising radiometric calibration.
\newblock {\em CVPR}, 2018.

\bibitem{Tozza17}
Silvia Tozza, William A.~P. Smith, Dizhong Zhu, Ravi Ramamoorthi, and Edwin~R.
  Hancock.
\newblock Linear differential constraints for photo-polarimetric height
  estimation.
\newblock {\em ICCV}, 2017.

\bibitem{Wolff97}
Lawrence~B. Wolff.
\newblock Polarization vision: {A} new sensory approach to image understanding.
\newblock {\em Image Vision Computing}, 1997.

\bibitem{xiong2014shading}
Ying Xiong, Ayan Chakrabarti, Ronen Basri, Steven~J Gortler, David~W Jacobs,
  and Todd Zickler.
\newblock From shading to local shape.
\newblock {\em IEEE transactions on pattern analysis and machine intelligence},
  37(1):67--79, 2014.

\bibitem{Yang18}
Luwei Yang, Feitong Tan, Ao Li, Zhaopeng Cui, Yasutaka Furukawa, and Ping Tan.
\newblock Polarimetric dense monocular {SLAM}.
\newblock {\em CVPR}, 2018.

\bibitem{ye2018single}
Wenjie Ye, Xiao Li, Yue Dong, Pieter Peers, and Xin Tong.
\newblock Single image surface appearance modeling with self-augmented cnns and
  inexact supervision.
\newblock {\em Wiley Online Library Computer Graphics Forum}, 2018.

\bibitem{Zhu_2019_CVPR}
Dizhong Zhu and William A.~P. Smith.
\newblock Depth from a polarisation + {RGB} stereo pair.
\newblock {\em CVPR}, 2019.

\end{thebibliography}
}

\end{document}